\documentclass{article}

\usepackage{microtype}
\usepackage{graphicx}
\usepackage{subcaption}
\usepackage{booktabs} 
\usepackage[table]{xcolor}

\usepackage{hyperref}

\usepackage[preprint]{icml2026}

\usepackage{amsmath}
\usepackage{amssymb}
\usepackage{mathtools}
\usepackage{amsthm}

\usepackage[capitalize,noabbrev]{cleveref}

\theoremstyle{plain}

\theoremstyle{definition}

\theoremstyle{remark}

\usepackage[textsize=tiny]{todonotes}

\icmltitlerunning{Rethinking Refinement: Correcting Generative Bias without Noise Injection}

\begin{document}

\twocolumn[
  \icmltitle{Rethinking Refinement: Correcting Generative Bias without Noise Injection}



  \icmlsetsymbol{equal}{*}

  \begin{icmlauthorlist}
    \icmlauthor{Xin Peng}{bupt,bupt2}
    \icmlauthor{Ang Gao}{bupt,bupt2}
  \end{icmlauthorlist}

\icmlaffiliation{bupt}{
State Key Laboratory of Information Photonics and Optical Communications, Beijing University of Posts and Telecommunications, Beijing 100876, China
}

\icmlaffiliation{bupt2}{
School of Physical Science and Technology, Beijing University of Posts and Telecommunications, Beijing 100876, China
}

\icmlcorrespondingauthor{Ang Gao}{anggao@bupt.edu.cn}

  \icmlkeywords{Machine Learning, ICML}

  \vskip 0.3in
]



\printAffiliationsAndNotice{}  

\begin{abstract}
Generative models, including diffusion and flow-based models, often exhibit systematic biases that degrade sample quality, particularly in high-dimensional settings.
We revisit refinement methods and show that effective bias correction can be achieved as a post-hoc procedure, without noise injection or multi-step resampling of the sampling process.
We propose a flow-matching–based \textbf{Bi-stage Flow Refinement (BFR)} framework with two refinement strategies operating at different stages: latent-space alignment for (approximately) invertible generators and data-space refinement trained with lightweight augmentations.
Unlike previous refiners that perturb sampling dynamics, BFR preserves the original ODE trajectory and applies deterministic corrections to generated samples.
Experiments on MNIST, CIFAR-10, and FFHQ 256$\times$256 demonstrate consistent improvements in fidelity and coverage; notably, starting from base samples with FID 3.95, latent-space refinement achieves a \textbf{State-of-the-Art} FID of \textbf{1.46} on MNIST using only a single additional function evaluation (1-NFE), while maintaining sample diversity.
\end{abstract}

\section{Introduction}

Generative models such as Denoising Diffusion Probabilistic Models (DDPMs)~\cite{sohl2015deep, ho2020denoising} and flow-matching-based continuous generative models~\cite{lipman2023flow, tong2024improving} have achieved remarkable success across diverse domains, including image synthesis, molecular conformer generation, and audio synthesis~\cite{dhariwal2021diffusion, nichol2021improved, rombach2021highresolution}. Unlike single-step generative mechanisms such as GANs~\cite{goodfellow2014generative} and VAEs~\cite{kingma2013auto}, diffusion and flow-matching models generate samples through iterative dynamics that transform simple base distributions (typically Gaussian noise) into complex target distributions. This iterative sampling process is central to their expressivity and stability.

A fundamental challenge in iterative generative frameworks lies in the mismatch between training and inference. During training, models learn to invert a forward noising or transformation process using ground-truth corrupted inputs at each step, whereas at inference they must condition on their own previously generated estimates. This discrepancy causes errors to accumulate progressively along the sampling trajectory, closely
resembling the exposure bias phenomenon studied in sequence modeling, where models are trained on ground-truth
sequences but evaluated on self-generated ones~\cite{bengio2015scheduled, schmidt2019generalization}. Recent studies in diffusion models have explicitly analyzed this training–inference gap, showing that it induces sampling drift and degrades sample quality~\cite{ning2023input}, and have proposed theoretical characterizations and mitigation strategies such as Epsilon Scaling~\cite{ning24_elucidating_exposure}.

In parallel, numerous refinement-based approaches have been proposed to enhance generative quality by introducing auxiliary or iterative refinement stages on intermediate or final outputs, instead of relying on a single-pass generation.
Many of these methods follow coarse-to-fine pipelines to recover high-frequency details or correct artifacts, ranging from early multi-scale models such as LAPGAN~\cite{denton2015lapgan} to more recent diffusion-based refinement frameworks, including iterative super-resolution and cascaded diffusion models~\cite{saharia2023image, ho2022cascaded}, diffusion--wavelet super-resolution~\cite{moser2024diffusion}, and diffusion-time--conditioned refiners~\cite{podell2024sdxl}.
Related approaches further refine preliminary samples via additional denoising or learned correction modules~\cite{pandey2022diffusevae}, or focus on restoration-oriented refinement such as deblurring and super-resolution~\cite{whang2022deblurring, saharia2023image}.
Another line of work repairs degraded outputs by injecting noise followed by denoising, as in Denoising Diffusion Restoration Models (DDRM)~\cite{kawar2022denoising} and Diffiner~\cite{sawata23_interspeech}; however, such noise-injection--based strategies necessarily revisit high-noise states during sampling, which may compromise stability.

\begin{figure}[t]
    \centering
    \includegraphics[width=0.75\linewidth]{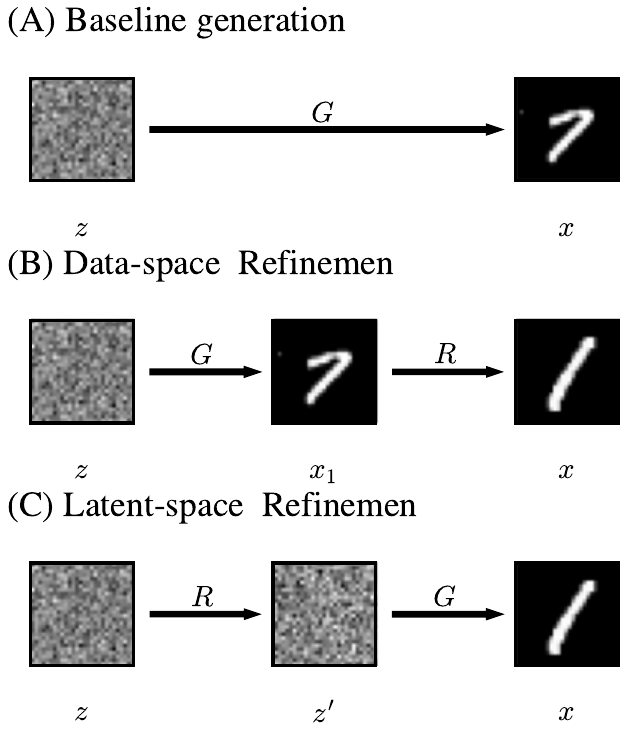}
    \caption{
    \textbf{Overview of bi-stage flow refinement (BFR).}
    (A) Baseline generation directly maps Gaussian noise $z$ to data samples $x$ via the base generator $G$.
    (B) \emph{Data-space refinement}: samples $x_1 = G(z)$ are further refined by a data-space flow $F$ to correct residual bias.
    (C) \emph{Latent-space refinement}: noise $z$ is first transformed into a bias-corrected latent $z'$ by a latent flow $F$, followed by generation through $G$.
    }
    \label{fig:bfr_overview}
\end{figure}

Despite these efforts, existing approaches either modify the training objective without directly addressing mapping biases induced by learned generative dynamics~\cite{xu2025flow}, or rely on inference-time perturbations that alter the sampling process~\cite{pandey2022diffusevae}.
We introduce \textbf{Bi-stage Flow Refinement (BFR)}, a principled approach for correcting generative bias in iterative models while preserving the original sampling dynamics.
Here, \emph{bi-stage} refers to refinement applied at two distinct stages of the generative process, yielding two independent strategies tailored to different model classes:

\begin{itemize}
    \item \textbf{Data-space Flow Refinement (DFR, Method 1).} 
    For non-invertible or general generative models, we refine preliminary generated samples through a flow-matching model trained on lightly augmented data (e.g., random blur or additive noise). Crucially, the augmentation is applied only during training, so that the original ODE-based sampling trajectory is preserved at inference time.
    
    \item \textbf{Latent-space Flow Refinement (LFR, Method 2).}
    For invertible generative models, we map real data to the latent space and observe that generative bias causes the recovered latent variables to deviate from the assumed Gaussian prior. 
    We train a flow-matching model to align the prior with the latent distribution induced by real data, enabling sampling from a bias-corrected latent space before generating data.

\end{itemize}

\section{Related Work}

\paragraph{Diffusion and Flow-Based Generative Models.}
Diffusion models and flow-based continuous generative models constitute two closely related families of iterative generative frameworks. 
Diffusion models learn to progressively denoise samples starting from Gaussian noise through a sequence of learned transitions~\cite{sohl2015deep, ho2020denoising, song2021scorebased}, while flow-matching-based models directly learn a continuous-time vector field that transports a simple prior distribution to the data distribution~\cite{lipman2023flow, tong2024improving}. 
Both paradigms rely on iterative dynamics and have demonstrated strong performance across images~\cite{dhariwal2021diffusion,nichol2021improved}, molecules~\cite{PhysRevD.100.034515,xu2022geodiff,wang2024generating}, and other modalities~\cite{kumar2019videoflow,Ki_2025_ICCV}.
However, their iterative nature also makes them sensitive to error accumulation and distribution mismatch between training and inference, which motivates the development of refinement strategies.

\paragraph{Refinement Methods.}
Refinement methods aim to improve the quality of samples produced by a base generative model~\cite{ho2022cascaded,whang2022deblurring, saharia2023image}, often via auxiliary correction stages on intermediate or final outputs~\cite{pandey2022diffusevae}. 
In molecular conformer generation, flow-matching-based refiners such as FMRefiner~\cite{xu2025flow} correct preliminary conformations by treating upstream outputs as noisy versions of true structures and training the refiner on artificially noised data. 
At inference, the learned flow refines the upstream output. 
While effective, these approaches have limitations: training and inference distributions are inconsistent, the amount of added noise is difficult to tune, and refiners do not explicitly align with underlying generative bias.

\paragraph{Discussion.}
Overall, existing refinement methods demonstrate the effectiveness of post-hoc correction for improving generative quality. 
At the same time, many of these methods rely on noise injection for additional sampling stages during inference, which alter the original deterministic ODE-based generation process and may introduce instability or increased computational cost~\cite{kawar2022denoising,sawata23_interspeech,pandey2022diffusevae}.
These limitations motivate the exploration of refinement strategies that improve robustness while preserving the original deterministic sampling dynamics.

\section{Background}

In this section, we introduce preliminaries for iterative generative models, taking denoising diffusion probabilistic models (DDPM) and flow matching (FM) as representative examples, and highlight the main sources of generative bias that arise during sample generation.

\subsection{Denoising Diffusion Probabilistic Models (DDPMs)}

DDPMs~\cite{sohl2015deep, ho2020denoising} define a forward diffusion process that gradually corrupts a data sample $x_0 \sim q(x_0)$ by adding Gaussian noise in $T$ steps:

\begin{equation}
    q(x_t \mid x_{t-1}) = \mathcal{N}\Big(x_t; \sqrt{1-\beta_t} x_{t-1}, \beta_t \mathbf{I}\Big), \quad t = 1,\dots,T,
\end{equation}

where $\beta_t$ is a variance schedule. The marginal distribution at step $t$ can be expressed as:

\begin{equation}
    q(x_t \mid x_0) = \mathcal{N}\Big(x_t; \sqrt{\bar{\alpha}_t} x_0, (1-\bar{\alpha}_t)\mathbf{I} \Big), \quad \bar{\alpha}_t = \prod_{s=1}^{t} (1-\beta_s).
\end{equation}

The generative (reverse) process approximates the inversion of the forward diffusion with a Gaussian transition, where the mean $\mu_\theta(x_t, t)$ is learned from data, and the covariance $\Sigma_\theta(x_t,t)$ can be either fixed according to a noise schedule~\cite{ho2020denoising} or learned~\cite{nichol2021improved}:

\begin{equation}
    p_\theta(x_{t-1} \mid x_t) = \mathcal{N}\big(x_{t-1}; \mu_\theta(x_t, t), \Sigma_\theta(x_t,t)\big).
\end{equation}

When the covariance $\Sigma_\theta$ is fixed~\cite{ho2020denoising}, the network predicts the mean $\mu_\theta(x_t,t)$, equivalently the noise $\epsilon_\theta(x_t,t)$, and is trained via
\begin{equation}
    \mathcal{L}_{\text{DDPM}}(\theta) = \mathbb{E}_{x_0, \epsilon, t} \Big[ \|\epsilon - \epsilon_\theta(x_t, t)\|^2 \Big],
\end{equation}
with $x_t = \sqrt{\bar{\alpha}_t} x_0 + \sqrt{1-\bar{\alpha}_t}\, \epsilon$, $\epsilon \sim \mathcal{N}(0,I)$.  
If the covariance is also learned, the full variational bound is optimized~\cite{nichol2021improved}.

At inference, samples are generated starting from $x_T \sim \mathcal{N}(0, \mathbf{I})$ by integrating a deterministic ordinary differential equation corresponding to the probability flow of the diffusion process~\cite{song2021denoising}:
\begin{equation}
\frac{d x_t}{d t}
=
-\frac{1}{2} \, \beta(t) \,\left( x_t
- \, \frac{\epsilon_\theta(x_t, t)}{\sqrt{1 - \bar{\alpha}_t}}\right) \, .
\end{equation}

\paragraph{Bias in DDPMs.}  
The discrepancy between training and inference—ground-truth noisy inputs $x_t$ vs. model-generated $\hat{x}_t$—leads to error accumulation, similar to exposure bias in autoregressive models~\cite{bengio2015scheduled}:

\begin{equation}
    \text{Bias} = \mathbb{E}\big[\hat{x}_{t-1} - x_{t-1} \mid \hat{x}_t\big].
\end{equation}

This bias grows over $T$ steps and degrades sample quality.

\subsection{Flow Matching for Generative Modeling}

Flow matching defines a continuous-time transport from a prior $p_0(x)$ to a target $p_1(x)$ via a vector field $v_\theta(x,t)$ satisfying
\begin{equation}
    \frac{\partial p_t(x)}{\partial t} + \nabla \cdot (p_t(x)\, v_\theta(x,t)) = 0,
\end{equation}

Given paired samples $(x_0, x_1)$, we consider a general interpolation path of the form
\begin{equation}
    x_t = a(t)\,x_0 + b(t)\,x_1, \quad t \in [0,1],
\end{equation}
where $a(t)$ and $b(t)$ are scalar interpolation functions satisfying
$a(0)=1$, $b(0)=0$, $a(1)=0$, and $b(1)=1$.
The corresponding target vector field is defined as
\begin{equation}
    f(x_t, t) = \dot{b}(t)\,x_1 + \dot{a}(t)\,x_0,
\end{equation}
which transports samples from $x_0$ to $x_1$ along the chosen path.

A common and simple choice is straight-line interpolation, given by
\begin{equation}
    x_t = (1-t)x_0 + t x_1,
    \quad
    f(x_t, t) = x_1 - x_0.
\end{equation}

and the training objective is
\begin{equation}
    \mathcal{L}_{\text{FM}}(\theta) = \mathbb{E}_{x_0,x_1,t}\big[\|v_\theta(x_t,t) - f(x_t,t)\|^2\big].
\end{equation}
\paragraph{Bias in FM.}  
Flow matching (FM) trains the network $v_\theta$ to predict the expected velocity over all interpolations between initial and terminal points at time $t$, rather than the instantaneous derivative $\dot{x}_t$ of a single trajectory. 
Formally, this systematic bias can be expressed as
\begin{equation}
    \text{Bias} = \mathbb{E}\big[v_\theta(x_t,t) - \dot{x}_t\big],
\end{equation}
where the expectation is taken over all interpolated paths.
In high-dimensional or multi-modal distributions, multiple divergent trajectories may share the same $x_t$ but have substantially different velocities, causing FM to average them into a biased vector field.
When integrated during sampling, this bias accumulates and leads samples to deviate from the ideal probability flow~\cite{liu2022flowstraightfastlearning, frans2024one}.

\subsection{ODE-Based Generative Dynamics}

Both diffusion and flow-matching generative models can be formulated as solving a deterministic probability flow ODE~\cite{song2021scorebased}:
\begin{equation}
    \frac{dx}{dt} = v(x,t),
\end{equation}
where $v(x,t)$ is the time-dependent velocity field induced by the learned model.

In this formulation, numerical integration over $t$ introduces discretization error. If $\hat{x}_t$ denotes the numerically integrated state,
\begin{equation}
    \hat{x}_t = x_0 + \int_0^t v(x_s,s)\,ds + \epsilon_{\text{int}},
\end{equation}
where $\epsilon_{\text{int}}$ is the integration (truncation) error due to solver discretization.

Exposure bias in diffusion models originates from the mismatch between training and inference trajectories, which can be analytically attributed to prediction errors at each sampling step.  
Numerical integration of the probability flow ODE introduces a secondary source of bias, $\epsilon_{\text{int}}$, which dominates only when using low-order solvers or large step sizes.  
When high-order solvers such as RK4 or Heun are employed, $\epsilon_{\text{int}}$ becomes negligible; consequently, methods like~\cite{ning24_elucidating_exposure} that primarily correct for integration-induced bias provide little improvement in this regime.

\subsection{Summary of Bias Sources}

Iterative generative models, including diffusion and flow-matching models, accumulate bias due to sequential transformations. The main sources are:

\begin{enumerate}
    \item Training-inference mismatch in DDPMs: network trained on ground-truth $x_t$ but inferring from model-generated $\hat{x}_t$.
    \item Expected velocity approximation in flow matching: network predicts $\mathbb{E}[\dot{x}_t]$ rather than the true instantaneous derivative.
    \item Numerical integration error in ODE solvers: truncation error $\epsilon_{\text{int}}$, which is secondary when using high-order solvers such as RK4 or Heun.
\end{enumerate}

\section{Method}
\label{sec:method}

We propose \textbf{Bi-stage Flow Refinement (BFR)}, a post-hoc framework to correct biases in pretrained generative models without retraining. 
Let $G_\theta : \mathcal{Z} \to \mathcal{X}$ be a generator with prior $p_0(z)=\mathcal{N}(0,I)$, inducing $q_\theta(x)=(G_\theta)_\sharp p_0$. 
BFR introduces a refinement operator $R$ such that the refined distribution approximates the data distribution:

\begin{equation}
    q_\theta^R =
    \begin{cases}
        (R \circ G_\theta)_\sharp p_0 \approx p_{\text{data}} & \text{(DFR)} \\[1mm]
        (G_\theta \circ R)_\sharp p_0 \approx p_{\text{data}} & \text{(LFR)}
    \end{cases}
\end{equation}

where $R$ operates in data space for general models (DFR) or latent space for invertible models (LFR).

All our refinement methods are formulated as ordinary differential equations (ODEs). 
The refinement operator $R$ is realized by integrating an ODE from the initial state to the refined state:
\begin{equation}
    u_1 = \text{ODE\_solve}(v_\phi, u_0, t\in[0,1]),
\end{equation}
where $v_\phi$ denotes the learned velocity field, $u_0$ is the input (either in data or latent space), 
and $\text{ODE\_solve}(\cdot)$ represents the numerical solver used to integrate the ODE from $t=0$ to $t=1$.

\subsection{Data-space Flow Refinement (DFR)}

Let
\begin{equation}
    \hat{x}_1 = G_\theta(z), \quad z \sim \mathcal{N}(0,I),
\end{equation}
be a preliminary sample generated by the base model, which generally deviates from the data distribution $x_0 \sim p_{\text{data}}$.
Data-space Flow Refinement (DFR) aims to learn a corrective vector field
$F_\psi(x,t)$ that transports $\hat{x}_1$ toward $x_0$ to reduce this bias.

Learning this transport directly can be unstable when $\hat{x}_1$ lies in low-density regions.
We therefore apply light data augmentation (e.g., random blur or additive noise)
\begin{equation}
    \tilde{x}_1 = \mathrm{DataAug}(\hat{x}_1),
\end{equation}
which slightly perturbs the sample while preserving its structure and stabilizes training.

Given $\tilde{x}_1$ and $x_0$, we define interpolation paths
\begin{equation}
    x_t = a(t)\tilde{x}_1 + b(t)x_0, \quad t\in[0,1].
\end{equation}
In practice, we use straight-line interpolation,
\begin{equation}
    x_t = (1-t)\tilde{x}_1 + t x_0,
    \quad
    f(x_t,t) = x_0 - \tilde{x}_1.
\end{equation}

The flow-matching objective is
\begin{equation}
    \mathcal{L}_{\text{DFR}}(\psi)
    =
    \mathbb{E}_{x_0,z,t}\!
    \big[ \| F_\psi(x_t,t) - f(x_t,t) \|^2 \big].
\end{equation}

At inference, refinement is performed deterministically by integrating
\begin{equation}
    \frac{d x(t)}{dt} = F_\psi(x(t),t),
    \quad x(0)=\hat{x}_1,\; x_{\text{refined}}=x(1).
\end{equation}

\begin{algorithm}[t]
\caption{DFR: Training Stage (Data-space Refinement)}
\begin{algorithmic}[1]
\STATE \textbf{INPUT:} Pretrained generator $G_\theta$, data distribution $p_{\text{data}}$, batch size $B$
\FOR{each training step}
    \STATE Sample real data $x_0 \sim p_{\text{data}}$ and latent $z \sim \mathcal{N}(0,I)$
    \STATE Generate preliminary sample $\hat{x}_1 = G_\theta(z)$
    \STATE Apply augmentation $\tilde{x}_1 = \mathrm{DataAug}(\hat{x}_1)$
    \STATE Sample $t \sim \text{Uniform}[0,1]$
    \STATE Interpolate $x_t = a(t)\,\tilde{x}_1 + b(t)\,x_0$
    \STATE Update $\psi$ via $\mathcal{L}_{\text{DFR}}$
\ENDFOR
\end{algorithmic}
\end{algorithm}

\begin{algorithm}[t]
\caption{DFR: Inference Stage (Data-space Refinement)}
\begin{algorithmic}[1]
\STATE \textbf{INPUT:} Pretrained generator $G_\theta$, trained refiner $F_\psi$, sample size $B$
\STATE Sample latent $z \sim \mathcal{N}(0,I)$
\STATE Generate preliminary sample $\hat{x}_1 = G_\theta(z)$
\STATE Refine sample $x = \text{ODE\_solve}(F_\psi, \hat{x}_1)$
\STATE \textbf{OUTPUT:} Refined samples $x$
\end{algorithmic}
\end{algorithm}

\paragraph{Existing refinement methods.}  
Existing data-space refiners have limitations: noise-injection approaches like DiffuseVAE~\cite{pandey2022diffusevae} require adding noise at sampling, while FMRefiner~\cite{xu2025flow} trains on artificially noised data that do not match upstream outputs and thus cannot fully correct generative bias (see Appendix~\ref{app:existing_refiners}).  
In contrast, our DFR uses lightweight data augmentations during training to improve robustness without noise injection at inference.
\begin{figure*}[t]
    \centering
    \includegraphics[width=0.95\textwidth]{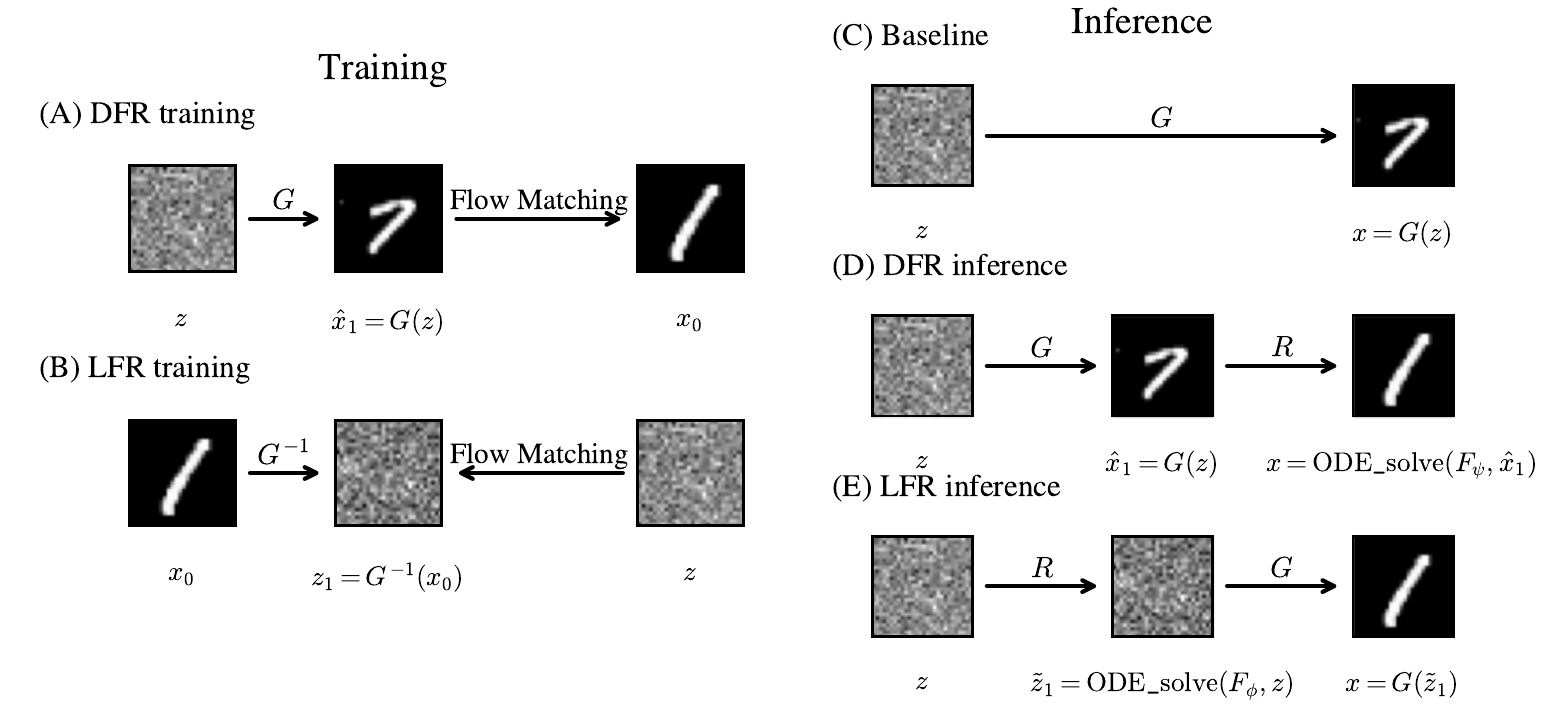}
    \caption{
    Illustration of Bi-stage Flow Refinement (BFR) for training and inference. 
    \textbf{Left:} Training stage, showing how DFR refines generated samples $\hat{x}_1$ 
    and LFR refines latent variables $z_1$. 
    \textbf{Right:} Inference stage, where the trained refiners $F_\psi$ (data-space) 
    and $F_\phi$ (latent-space) are applied via ODE solvers to produce corrected samples.
    }
    \label{fig:bfr_training_inference}
\end{figure*}

\subsection{Latent-space Flow Refinement (LFR)}

For an invertible generator $G_\theta$, data is mapped as
\begin{equation}
    z_1 = G_\theta^{-1}(x_0),
\end{equation}
where for DDPM and FM, inversion is done by reversing the probability flow ODE.
Ideally, the inverse-mapped latents follow the prior $z_1 \sim
\mathcal{N}(0,I)$, but generative bias distorts this distribution, motivating a
latent-space refinement model to align data-induced latents with the Gaussian prior.

We therefore train a flow-matching model $F_\phi(z,t)$ to map standard Gaussian
noise to the biased latent distribution induced by $G_\theta^{-1}$.
Directly regressing from $z$ to $z_1$, however, is suboptimal due to two
unavoidable sources of error.

\paragraph{Reconstruction error.}
Practical implementations of generative models may fail to perfectly reconstruct samples. 
For instance, latent-space compression in VAEs~\cite{kingma2013auto} or in diffusion models that operate in a pre-trained latent space (e.g., Latent Diffusion~\cite{rombach2021highresolution}) introduces irreversibility, 
and numerical errors can further produce deviations~\cite{pmlr-v130-behrmann21a}:
\begin{equation}
    \varepsilon_{\text{rec}} = x_0 - G_\theta(z_1) \neq 0,
\end{equation}
meaning the latent code $z_1$ does not correspond to an exact reconstruction of $x_0$.

\paragraph{Refinement error.} 
The refinement network $F_\phi$ maps the initial latent $z_0$ to a predicted terminal state $\hat{z}_1$. 
Ideally, $\hat{z}_1$ should match the true terminal latent $z_1$. 
In practice, however, imperfect optimization, and numerical integration errors cause $\hat{z}_1$ to deviate from $z_1$, resulting in the refinement error
\begin{equation}
\varepsilon_{\text{ref}} = \hat{z}_1 - z_1.
\end{equation}

\paragraph{Latent noise mixing.}
To jointly account for both errors and improve robustness, we perturb the target
latent with a small Gaussian component:
\begin{equation}
    z_1^\alpha
    = \sqrt{1-\alpha^2}\, z_1 + \alpha z,
    \quad z \sim \mathcal{N}(0,I), \; \alpha \ll 1.
\end{equation}
Here, $\alpha$ can be implemented either as a fixed constant or as a randomly
sampled mixing coefficient within a predefined range, with both variants serving
the same role of controlling the magnitude of latent noise perturbation.
This noise mixing regularizes the refinement target, preventing overfitting to
potentially inaccurate inverse latents and promoting controlled latent-space
exploration. Similar stabilizing effects of stochastic perturbations have been
observed in noise-based strategies for reinforcement learning and latent
optimization~\cite{plappert2018parameter, mahankali2024random, zhao2025simflow}.
However, excessively large $\alpha$ can overly deviate from $z_1$ and compromise
effective refinement; practical guidelines for choosing $\alpha$ are provided in
Appendix~\ref{app:alpha}.

The flow-matching interpolation and target are defined as
\begin{equation}
\begin{aligned}
    x_t &= a(t)\, z + b(t)\, z_1^\alpha, \\
    f(z,t) &= \dot{a}(t)\, z + \dot{b}(t)\, z_1^\alpha,
    \qquad t \in [0,1].
\end{aligned}
\end{equation}

with $a(0)=1, b(0)=0$ and $a(1)=0, b(1)=1$, and training objective
\begin{equation}
    \mathcal{L}_{\text{LFR}}(\phi)
    = \mathbb{E}_{z,x_t,t}
    \big[ \| F_\phi(x_t, t) - f(z,t) \|^2 \big].
\end{equation}

\paragraph{Inference.}
At inference time, the latent refiner is applied deterministically to samples drawn
from the prior.
Specifically, we sample
\begin{equation}
    z \sim \mathcal{N}(0,I),
\end{equation}
and obtain the refined latent by solving the corresponding probability flow ODE:
\begin{equation}
    \tilde{z}_1 = \text{ODE\_solve}(F_\phi, z, t \in [0,1]).
\end{equation}
The final data sample is then generated via the base generator:
\begin{equation}
    x = G_\theta(\tilde{z}_1).
\end{equation}

Unlike the training phase, where a small amount of Gaussian perturbation is introduced
to improve robustness against reconstruction and refinement errors, inference is
performed without any additional noise mixing. This preserves the deterministic
sampling dynamics of the underlying flow and retains the original generative capacity
of the base model.

\paragraph{Latent- vs. data-space refinement.}
We distinguish refinement strategies by the space in which bias correction is performed.
Data-space refinement (DFR) operates directly on generated samples and is therefore constrained once a sample is fixed.
In contrast, latent-space flow refinement (LFR) corrects bias at its source by directly regularizing the latent distribution,
enforcing
\begin{equation}
    G_\theta^{-1}(x_0) \xrightarrow{\text{LFR}} \mathcal{N}(0,I),
\end{equation}
thereby aligning data-induced latents with the prescribed Gaussian prior.

\paragraph{Relation to noise optimization and RL.}
In contrast to latent-space flow refinement (LFR), RL-based latent optimization methods~\cite{eyring2025noise, zhou2025golden, li2025enhancing}
seek a single optimized latent code by solving
\begin{equation}
    z^\star = \arg\max_z \; \mathbb{E}\big[ \mathcal{J}(G_\theta(z)) \big],
\end{equation}
where $\mathcal{J}(\cdot)$ denotes an external reward function.
Such methods rely on reward design, seed sensitivity, and per-sample optimization.
In contrast, LFR requires neither reward signals nor iterative latent search;
it directly reshapes the latent statistics, enabling unbiased and efficient generation.

\begin{algorithm}[t]
\caption{LFR: Training Stage (Latent-space Refinement)}
\begin{algorithmic}[1]
\STATE \textbf{INPUT:} Pretrained invertible generator $G_\theta$, data distribution $p_{\text{data}}$, batch size $B$, mixing coefficient $\alpha \ll 1$
\FOR{each training step}
    \STATE Sample real data $x_0 \sim p_{\text{data}}$ and latent $z,z_0 \sim \mathcal{N}(0,I)$
    \STATE Compute $z_1 = G_\theta^{-1}(x_0)$
    \STATE Perturb latent: $z_1^\alpha = \sqrt{1-\alpha^2}\, z_1 + \alpha z_0$
    \STATE Sample $t \sim \text{Uniform}[0,1]$
    \STATE Interpolate: $x_t = a(t)\, z + b(t)\, z_1^\alpha$
    \STATE Update $\phi$ via $\mathcal{L}_{\text{LFR}}$
\ENDFOR
\end{algorithmic}
\end{algorithm}

\begin{algorithm}[t]
\caption{LFR: Inference Stage (Latent-space Refinement)}
\begin{algorithmic}[1]
\STATE \textbf{INPUT:} Trained latent refiner $F_\phi$, invertible generator $G_\theta$, sample size $B$
\STATE Sample latent $z \sim \mathcal{N}(0,I)$
\STATE Refine latent via ODE: $\tilde{z}_1 = \text{ODE\_solve}(F_\phi, z)$
\STATE Generate final sample: $x = G_\theta(\tilde{z}_1)$
\STATE \textbf{OUTPUT:} Refined samples $x$
\end{algorithmic}
\end{algorithm}

\section{Experiments \& Results}

We evaluate our \textbf{Bi-stage Flow Refinement (BFR)} framework on image and molecular generation tasks, comparing base models, BFR refiners, and representative existing data-space refiners (details in Appendix~\ref{app:existing_refiners}).Compared methods include:  
\begin{itemize}
  \item \textbf{Base models}: DDPM~\cite{ho2020denoising}, FM~\cite{lipman2023flow}.
  \item \textbf{Base + Existing Refiners}: DiffuseVAE~\cite{pandey2022diffusevae}, FMRefiner~\cite{xu2025flow}. 
  \item \textbf{Base + BFR refiners}:
    \begin{itemize}
        \item \textbf{DFR}: Data-space refinement on lightly augmented data; augmentation applied only during training.
        \item \textbf{LFR}: Latent-space refinement for invertible models, aligning data-induced latent distributions with the Gaussian prior.
    \end{itemize}
\end{itemize}

We conduct experiments on a single NVIDIA RTX 4090 GPU. Image benchmarks include MNIST~\cite{lecun1998gradient}, CIFAR-10~\cite{krizhevsky2009learning}, and FFHQ 256$\times$256~\cite{karras2019style}, using a standard U-Net architecture~\cite{ronneberger2015u,ho2020denoising}. 
Molecular generation tasks include ALA2~\cite{pande2003atomistic} and Chignolin Mutant\footnote{\url{http://ftp.mi.fu-berlin.de/pub/cmbdata/bgmol/datasets/chignolin/ChignolinOBC2PT.tgz}}, following the molecular flow models in Flow Perturbation~\cite{peng2025flow}. Detailed molecular results are reported in Appendix~\ref{app:molecular}.

\subsection{Results: Image Generation}

Table~\ref{tab:image_results_twocol_metrics} summarizes image generation performance on CIFAR-10, MNIST, and FFHQ 256$\times$256. We report FID~\cite{heusel2017gans}, sFID~\cite{szegedy2016rethinking}, and IS~\cite{salimans2016improved}, where lower FID/sFID and higher IS indicate better sample quality.

Across datasets, both variants of our Bi-stage Flow Refinement (BFR) consistently improve over the base generative models. 
On MNIST, a single-step latent-space refinement (LFR, 1-NFE) reduces FID from 3.95 to 1.46, 
far surpassing the best previously reported FID of 4.5 for direct generative models~\cite{dai2021slicediterativenormalizingflows}. 
Here, single-step (1-NFE) indicates that the refinement ODE is solved with a single function evaluation, 
i.e., the numerical solver integrates the learned velocity field with only one step. 
This demonstrates that lightweight post-hoc refinement can dramatically improve sample quality 
with minimal additional cost, while preserving diversity, as reflected by stable IS scores.

Interestingly, increasing the number of NFEs for LFR from 1 to 10 consistently degrades sample quality across all datasets. 
This over-refinement occurs because additional steps can push latent variables too far, overshooting the optimal correction. 
In contrast, DFR can benefit from more NFEs, as iterative refinement in data space can progressively correct generative bias, improving quality on some datasets.

On CIFAR-10, MNIST, and FFHQ 256$\times$256, both data-space and latent-space refinements consistently improve FID and sFID over the base models.
In particular, LFR achieves the best overall performance across datasets.
Full quantitative results are summarized in Table~\ref{tab:image_results_twocol_metrics}.

\begin{figure}[t]
    \centering
    \includegraphics[width=\linewidth]{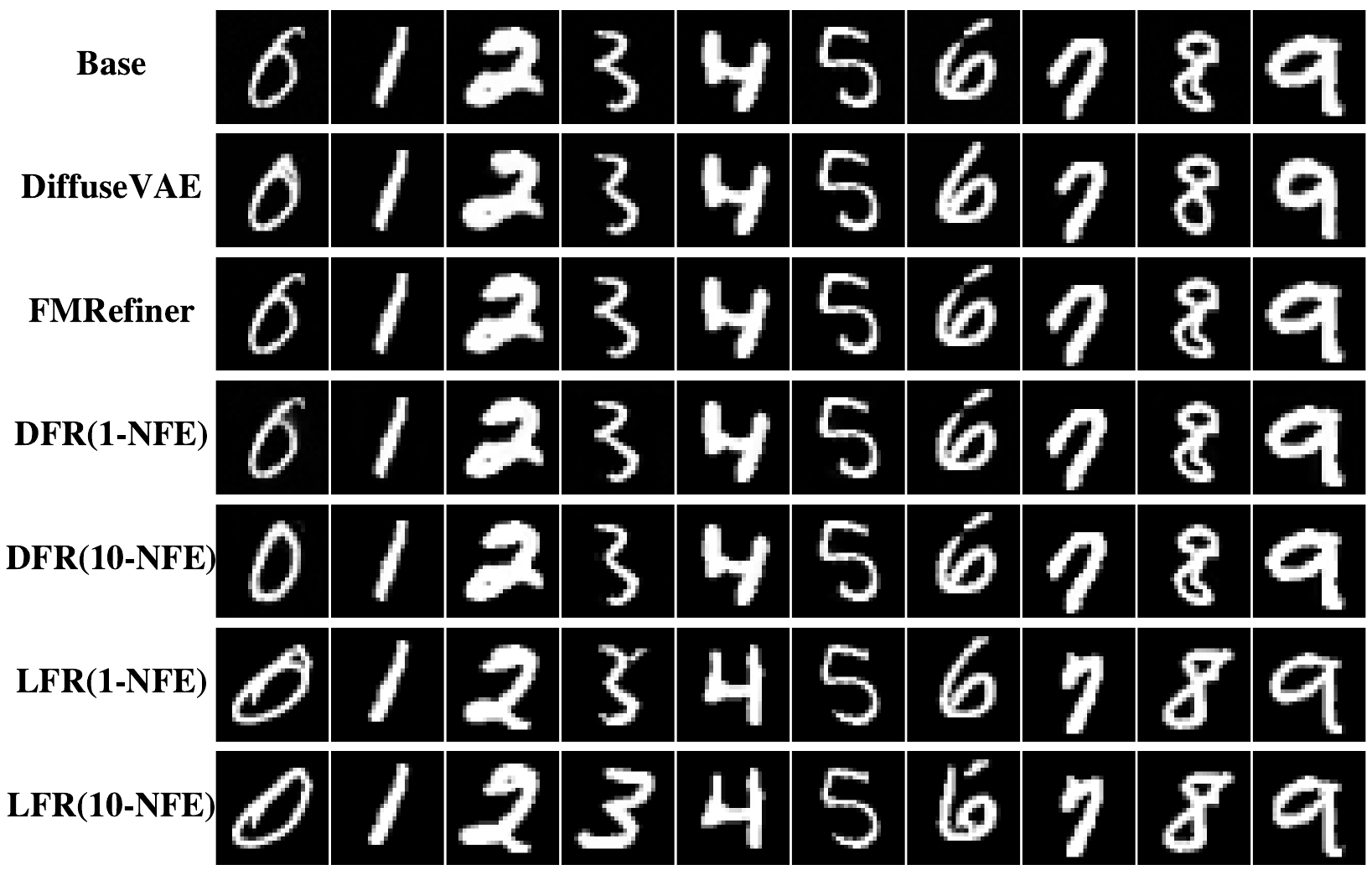}
    \caption{Qualitative comparison of MNIST samples before and after refinement.
    Each row corresponds to a refinement method. DFR and LFR significantly improve visual quality.}
    \label{fig:mnist_refinement_qualitative}
\end{figure}

Figure~\ref{fig:mnist_refinement_qualitative} provides qualitative examples of MNIST refinement. Both DFR and LFR sharpen digit structures and suppress spurious artifacts, with latent-space refinement achieving comparable or better visual quality using significantly fewer function evaluations.

Overall, these results confirm that BFR effectively mitigates generative bias and improves sample quality, with latent-space refinement providing the most favorable trade-off between performance and computational cost.

\begin{figure}[t]
    \centering
    \includegraphics[width=\linewidth]{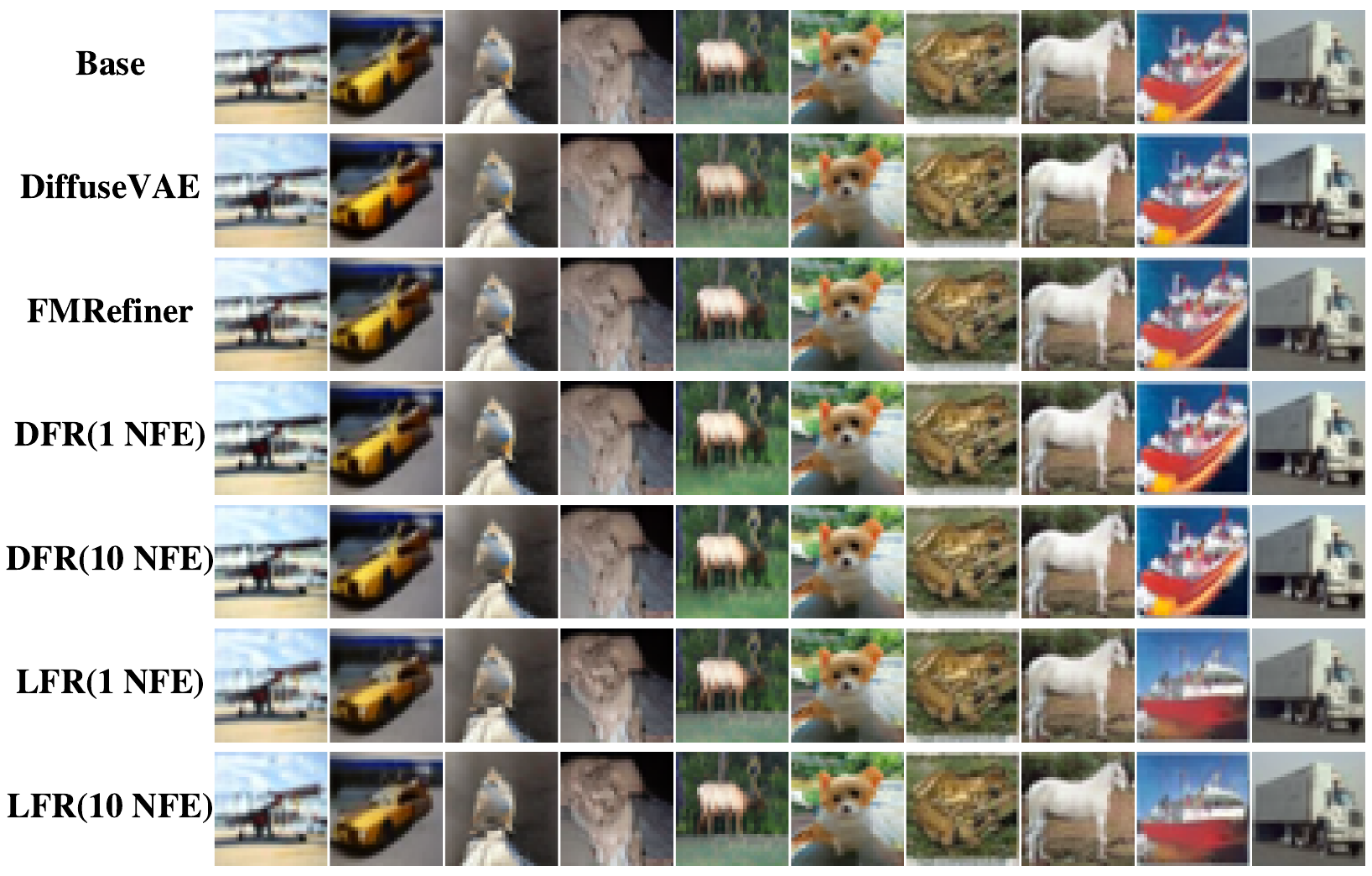}
    \caption{Qualitative comparison of CIFAR-10 samples before and after refinement.
    Each row corresponds to a refinement method.
    DFR and LFR improve visual fidelity over the base model.}
    \label{fig:cifar_refinement_qualitative}
\end{figure}

\begin{figure}[t]
    \centering
    \includegraphics[width=\linewidth]{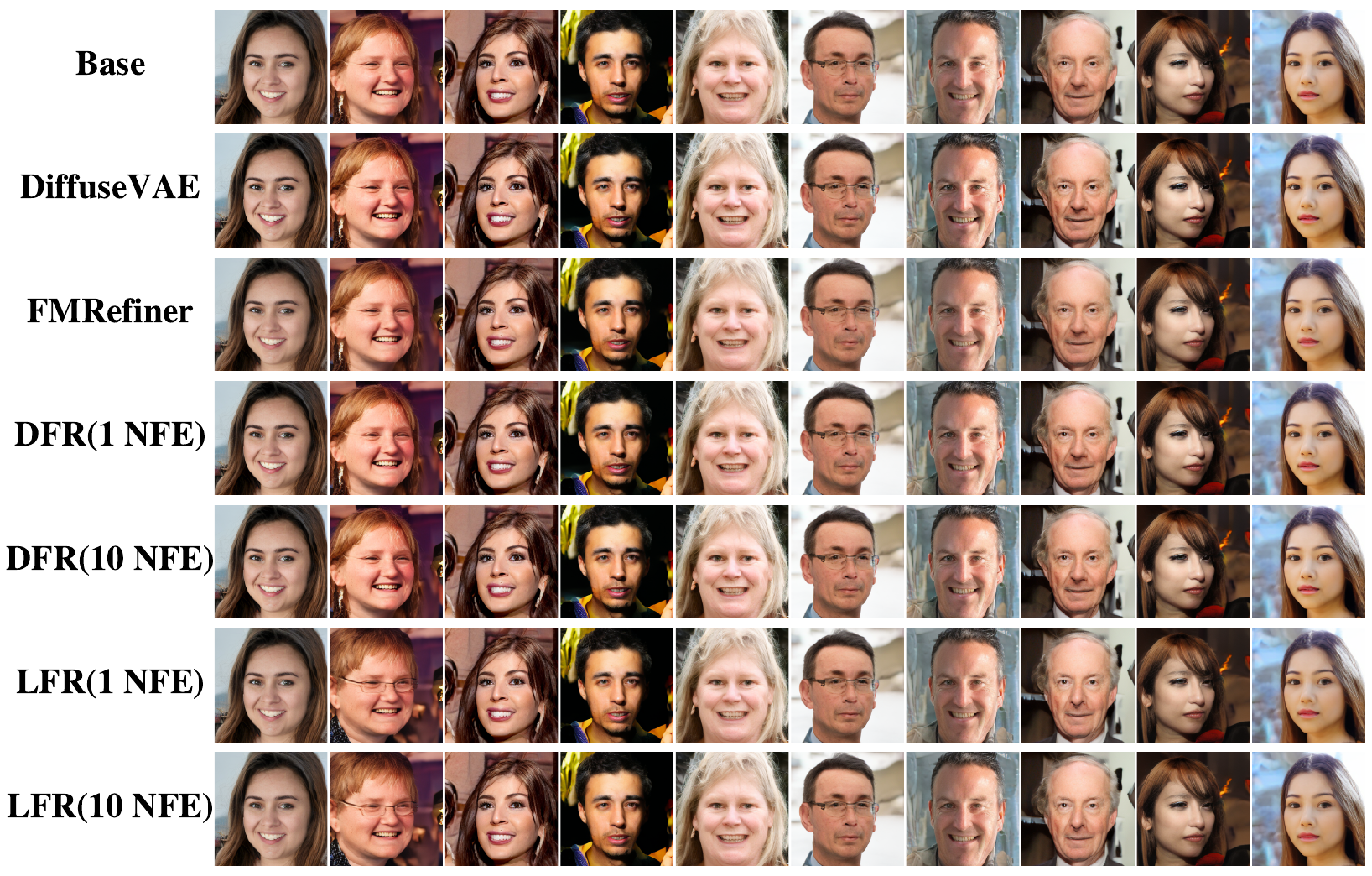}
    \caption{Qualitative comparison of FFHQ 256$\times$256 samples before and after refinement.
    Each row corresponds to a refinement method.
    DFR and LFR improve visual fidelity over the base model.}
    \label{fig:FFHQ_refinement_qualitative}
\end{figure}

\begin{table*}[t]
\centering
\caption{Image generation results on CIFAR-10, MNIST, and FFHQ 256$\times$256 
(FID $\downarrow$ / sFID $\downarrow$ / IS $\uparrow$). 
BFR consistently improves generation quality over base models, with latent-space refinement achieving strong performance under limited sampling steps.}
\label{tab:image_results_twocol_metrics}
\small
\begin{tabular}{lccccccccc}
\toprule
Model & \multicolumn{3}{c}{CIFAR-10} & \multicolumn{3}{c}{MNIST} & \multicolumn{3}{c}{FFHQ 256$\times$256} \\
\cmidrule(lr){2-4} \cmidrule(lr){5-7} \cmidrule(lr){8-10}
 & FID $\downarrow$ & sFID $\downarrow$ & IS $\uparrow$
 & FID $\downarrow$ & sFID $\downarrow$ & IS $\uparrow$
 & FID $\downarrow$ & sFID $\downarrow$ & IS $\uparrow$ \\
\midrule
Base & 4.84 & 0.0034 & 9.04 & 3.95 & 0.0036 & 2.077 & 9.47 & 0.0049 & 4.44 \\
+ DiffuseVAE~\cite{pandey2022diffusevae}  &4.42 & 0.0015 &\cellcolor{gray!15}\textbf{9.70} & 3.92 & 0.0028 & 2.093&9.32 & 0.0054 &4.48 \\
+ FMRefiner~\cite{xu2025flow}  &4.43 &   0.0026  &  9.29 & 2.87 &  0.0023 & 2.075& 9.27 & 0.0047 & 4.48\\
\midrule
+ DFR (1-NFE) & 3.96 & 0.0020 & 9.31 & 2.42 & 0.0015 & \cellcolor{gray!15}\textbf{2.139} & 9.04 & 0.0049 & \cellcolor{gray!15}\textbf{4.54} \\
+ DFR (10-NFE) & 3.94 & 0.0019 & 9.40 & 2.19 & 0.0015 & 2.123 & 8.99 & 0.0048 & 4.51 \\
+ LFR (1-NFE) &  \cellcolor{gray!15}\textbf{3.10} & \cellcolor{gray!15}\textbf{0.0014} & 9.41 & \cellcolor{gray!15}\textbf{1.46} & \cellcolor{gray!15}\textbf{0.0008} & 2.085 & \cellcolor{gray!15}\textbf{8.75} & 0.0041 & 4.41 \\
+ LFR (10-NFE) & 3.13 & 0.0015 & 9.40 & 2.18 & 0.0016 & 2.095 & 8.78 & \cellcolor{gray!15}\textbf{0.0040} & 4.39 \\
\bottomrule
\end{tabular}
\end{table*}

\subsection{Robustness to Base Model Shift}

We examine whether BFR refiners generalize across different instances of the same base model.
On MNIST, we consider two generators with identical architecture and training objective, 
but trained for different numbers of iterations, resulting in different training quality
(FID 3.95 vs.\ 7.61).
Refiners trained on the higher-quality generator remain effective when applied to the lower-quality one without retraining: DFR reduces the FID to 4.62, while LFR reduces it to 6.86.

This transferability stems from the design of our refinement schemes.
DFR uses data augmentation to capture systematic data-level biases, while LFR employs latent-space noise mixing to enhance robustness across model instances.
Consequently, both refiners correct shared generative biases, with data-space refinement being more effective in this setting.

This robustness enables reuse of a single refiner across generators of varying quality within the same model family, reducing training and deployment costs.
Additional results are provided in Appendix~\ref{Robustness}.

\section{Discussion and Analysis}

Data-space refinement corrects visible artifacts directly on generated samples, enabling strong cross-model generalization, but its training is tightly coupled to specific sampling dynamics or noise schedules, limiting flexibility. In contrast, latent-space refinement offers greater modeling flexibility under an invertible base generator. From a unified perspective, prior works—including latent diffusion models~\cite{rombach2021highresolution}, latent normalizing flows like STARFlow~\cite{gu2025starflow} and SimFlow~\cite{zhao2025simflow}, and latent-space flow matching~\cite{dao2023flow}—can be interpreted as forms of latent-space refinement: first learning a latent representation (typically via a VAE~\cite{kingma2013auto}) that emphasizes reconstruction fidelity, then applying a secondary model to restore distributional alignment~\cite{higgins2017betavae,burgess2018understanding, shao2020dynamicvae}.

Importantly, our results show that even when base generators produce high-quality samples, 
and latent-space generative modeling (e.g., VAE with flow matching) is already applied, 
an additional latent-space refinement step (LFR) still improves sample quality. 
Here, the latent space refined by LFR is not the encoder latent space of a VAE, 
but the operational latent space corresponding to the model's initial sampling distribution (i.e., the noise or prior space $p_0(z)$).

\section{Conclusion}

We propose \textbf{Bi-stage Flow Refinement (BFR)}, a general framework for post-hoc generative refinement in both data and latent spaces. BFR effectively mitigates systematic bias in pretrained models, improving sample quality across image and molecular generation tasks. Data-space refinement directly corrects visible artifacts and exhibits strong cross-model transferability, while latent-space refinement provides flexible, low-cost improvements, achieving state-of-the-art FID on MNIST with a single function evaluation (1-NFE) without sacrificing diversity.

Our analysis further reveals the generality of latent-space refinement: many existing methods first generate a latent representation, 
often with a compressed dimensionality, and then apply a secondary model to correct residual errors. 
Crucially, we show that LFR remains effective even when no explicit latent-space compression mechanism (such as in latent-space diffusion or flow matching) is applied, 
demonstrating that refinement in the latent space can significantly improve sample quality regardless of the dimensionality constraints of the initial representation.

\section*{Acknowledgments}
This work was supported by the Opening Project of the State Key Laboratory of Information Photonics and Optical Communications (Grant No. IPOC2025ZT02), and by the National Natural Science Foundation of China (Grant No. 22473016).

\section*{Software and Data}

To support reproducibility, we provide an anonymous implementation of all methods
described in this paper, along with scripts for data generation and evaluation.
The code is publicly available at the following anonymous repository:
\begin{center}
\url{https://github.com/XinPeng76/Rethinking-Refinement}
\end{center}

\section*{Impact Statement}

This work proposes Bi-stage Flow Refinement (BFR), a post-hoc framework to improve the fidelity and stability of samples generated by existing generative models. 
The primary goal of our research is to advance the field of generative modeling by providing more reliable and accurate sample refinement techniques. 

While our methods are intended for research and applications in image and molecular generation, they could, in principle, be applied to synthetic media generation more broadly. 
Potential ethical considerations include the misuse of improved generative models for creating deceptive content. 
We emphasize that BFR itself does not introduce new data sources or biases beyond what exists in the base generative models, and it can help researchers better understand and mitigate generative biases in existing models.

Overall, we believe that the societal impact of this work is largely positive, as it promotes safer, more accurate, and more controllable generative modeling, while potential misuse risks are similar to those of standard generative models.

\nocite{langley00}

\bibliography{example_paper}
\bibliographystyle{icml2026}

\newpage
\appendix
\onecolumn
\section{Existing Data-space Refinement Methods}
\label{app:existing_refiners}

This appendix reviews two representative data-space refinement methods that are commonly
used to improve the quality of intermediate samples produced by generative models.
We summarize their training and inference procedures, discuss their inherent limitations,
and provide additional empirical comparisons on MNIST.
These observations motivate the design of our Data-space Flow Refinement (DFR).

Both methods attempt to correct errors in generated samples directly in data space.
However, they either require modifying the sampling trajectory at inference
or rely on synthetic noise assumptions that do not match the true generative bias.

\subsection{Noise-injection Refiners (DiffuseVAE)}
\label{app:diffusevae}

Noise-injection refiners assume that injecting small perturbations into intermediate
samples allows a separate refiner network to recover higher-quality outputs.
Let $x_0$ denote a clean target sample and $\hat{x}_1 = G_\theta(z)$ a preliminary sample generated by the upstream model.
A perturbed version is constructed as
\begin{equation}
    x_1' = \hat{x}_1 + \sigma_d \epsilon, \quad \epsilon \sim \mathcal{N}(0,I),
\end{equation}
where $\sigma$ controls the noise magnitude.

In the original DiffuseVAE~\cite{pandey2022diffusevae} formulation, the refiner is trained using a DDPM-style interpolation:
\begin{equation}
    x_t = a(t) x_0 + b(t) x_1', \quad t \in [0,1],
\end{equation}
where $a(t), b(t)$ follow the DDPM noise schedule.
In our flow-matching experiments, we adopt a simpler linear interpolation:
\begin{equation}
    x_t = t x_0 + (1-t) x_1',
\end{equation}
to construct intermediate points for refiner training.

The target vector field is then defined as $v_t = x_0 - x_1'$, and the refiner is trained via
\begin{equation}
    \mathcal{L}_{\text{NI}}(\psi)
    = \mathbb{E}_{z,\epsilon,t}\!\left[
    \| R_\psi(x_t, t) - v_t \|^2
    \right],
\end{equation}
which is closely related to diffusion and flow-matching objectives.

\begin{minipage}{0.48\textwidth}
\begin{algorithm}[H]
\caption{Noise-injection Refiner: Training}
\begin{algorithmic}[1]
\STATE Sample $z \sim \mathcal{N}(0,I)$
\STATE Generate $\hat{x}_1 = G_\theta(z)$
\STATE Add noise: $x_1' = \hat{x}_1 + \sigma_d\epsilon, \quad \epsilon \sim \mathcal{N}(0,I)$
\STATE Sample $t \sim \mathcal{U}[0,1]$, form $x_t = t x_0 + (1-t) x_1'$
\STATE Update $R_\psi$ using $\|R_\psi(x_t,t) - (x_0 - x_1')\|^2$
\end{algorithmic}
\end{algorithm}
\end{minipage}\hfill
\begin{minipage}{0.48\textwidth}
\begin{algorithm}[H]
\caption{Noise-injection Refiner: Inference}
\begin{algorithmic}[1]
\STATE Generate $\hat{x}_1 = G_\theta(z), \quad z \sim \mathcal{N}(0,I)$
\STATE Add noise: $x_1' = \hat{x}_1 + \sigma_d\epsilon, \quad \epsilon \sim \mathcal{N}(0,I)$
\STATE Refine via ODE solving or denoising: $x = \text{ODE\_solve}(R_\psi, x_1')$
\end{algorithmic}
\end{algorithm}
\end{minipage}

\paragraph{Hyperparameter tuning (DiffuseVAE).}
For DiffuseVAE, the noise magnitude $\sigma$ is treated as a tunable hyperparameter.
We perform a grid search over a small set of candidate values
($\sigma_d \in \{0.01, 0.05, 0.1, 0.2\}$, depending on the dataset),
and select the near-optimal $\sigma_d$ based on validation FID.
The results reported in Tables~\ref{tab:image_results_twocol_metrics} use the best-performing $\sigma_d$
for DiffuseVAE on each dataset.
Table~\ref{tab:sigma_ablation} reports a representative ablation study,
illustrating the sensitiv

\paragraph{Limitations.}
\begin{itemize}
    \item The inference trajectory must be explicitly modified by injecting noise,
    breaking the original deterministic ODE path.
    
    \item Performance is sensitive to the choice of noise magnitude $\sigma_d$,
    which requires careful tuning and may introduce artifacts. In our experiments,
    we use a near-optimal value of $\sigma_d$ selected based on preliminary ablations
    to ensure stable and high-quality refinement.

\end{itemize}

\begin{table*}[t]
\centering
\caption{
Ablation study on noise magnitude $\sigma_d$ for DiffuseVAE refiners.
Results are reported on CIFAR-10, MNIST, and FFHQ 256$\times$256.
FID $\downarrow$ / sFID $\downarrow$ / IS $\uparrow$ are reported.
}
\label{tab:sigma_ablation}
\small
\begin{tabular}{lccccccccc}
\toprule
$\sigma_d$
& \multicolumn{3}{c}{CIFAR-10}
& \multicolumn{3}{c}{MNIST}
& \multicolumn{3}{c}{FFHQ 256$\times$256} \\
\cmidrule(lr){2-4} \cmidrule(lr){5-7} \cmidrule(lr){8-10}
& FID $\downarrow$ & sFID $\downarrow$ & IS $\uparrow$
& FID $\downarrow$ & sFID $\downarrow$ & IS $\uparrow$
& FID $\downarrow$ & sFID $\downarrow$ & IS $\uparrow$ \\
\midrule
Base
& 4.84 & 0.0034 & 9.04
& 3.95 & 0.0036 & 2.077
&9.47 & \cellcolor{gray!15}\textbf{0.0049} & 4.44 \\
0.01
& \cellcolor{gray!15}\textbf{4.42} & \cellcolor{gray!15}\textbf{0.0015} & 9.70
& 5.68 & 0.0051 & 2.110
& \cellcolor{gray!15}\textbf{9.32} & 0.0054 &\cellcolor{gray!15}\textbf{4.48} \\
0.05
& 11.60 & 0.0086 & \cellcolor{gray!15}\textbf{9.73}
& 14.38 & 0.0135 & \cellcolor{gray!15}\textbf{2.117}
& 10.03 & 0.0058 & 4.53 \\
0.10
& 15.22 & 0.0107 & 9.09
& \cellcolor{gray!15}\textbf{3.92} & \cellcolor{gray!15}\textbf{0.0028} & 2.093
& 13.59 & 0.0092 & 4.53 \\
0.20
& 19.90 & 0.0134 & 8.15
& 5.01 & 0.0041 & 2.103
& 31.65 & 0.0281 & 4.28 \\
\bottomrule
\end{tabular}
\end{table*}

\subsection{Flow-matching--based Refiner (FMRefiner)}
\label{app:fmrefiner}

FMRefiner assumes that the preliminary sample resembles a noisy version of real data.
Training relies on synthetic pairs $(x_0, x_1)$ constructed as
\begin{equation}
    x_1 = x_0 + \sigma_f \epsilon, \quad \epsilon \sim \mathcal{N}(0,I),
\end{equation}
where $\sigma_f$ controls the perturbation strength.
Random interpolations are defined as
\begin{equation}
    x_t = \alpha(t)x_0 + \beta(t)x_1 + s(t)z,
    \quad z \sim \mathcal{N}(0,I),
\end{equation}
where $\alpha(t), \beta(t)$ and $s(t)$ are predefined scalar functions.
In our main experiments, for consistency across refinement methods, we adopt a simple symmetric parameterization
\begin{equation}
    x_t = t x_0 + (1 - t) x_1 + \sigma_z\, t(1 - t) z,
\end{equation}
which injects noise primarily at intermediate time steps.

A refiner $R_\psi$ is then trained to map $x_1$ back to $x_0$.

\begin{minipage}{0.48\textwidth}
\begin{algorithm}[H]
\caption{FMRefiner: Training}
\begin{algorithmic}[1]
\STATE Sample $x_0 \sim p_{\text{data}}$, $z \sim \mathcal{N}(0,I)$
\STATE Construct $x_1 = x_0 + \sigma_f\epsilon, \quad \epsilon \sim \mathcal{N}(0,I)$
\STATE Form interpolation $x_t = \alpha(t)x_0 + \beta(t)x_1 + s(t)z$
\STATE Update $R_\psi$ to predict $x_0$
\end{algorithmic}
\end{algorithm}
\end{minipage}\hfill
\begin{minipage}{0.48\textwidth}
\begin{algorithm}[H]
\caption{FMRefiner: Inference}
\begin{algorithmic}[1]
\STATE Generate $\hat{x}_1 = G_\theta(z), \quad z \sim \mathcal{N}(0,I)$
\STATE Apply refiner: $x = \text{ODE\_solve}(R_\psi, \hat{x}_1)$
\end{algorithmic}
\end{algorithm}
\end{minipage}

\paragraph{Hyperparameter tuning (FMRefiner).}
For FMRefiner, the perturbation magnitude $\sigma_f$ in the construction of $x_1$
is treated as a tunable hyperparameter, while the intermediate noise scale
$\sigma_z$ in the interpolation is fixed to $0.1$ across all experiments.
We perform a grid search over a small set of candidate values of $\sigma_f$
($\sigma_f \in \{0.01, 0.05, 0.1, 0.2\}$, depending on the dataset),
and select the near-optimal $\sigma_f$ based on validation FID.
The results reported in Tables~\ref{tab:image_results_twocol_metrics} correspond to the best-performing
$\sigma_f$ for FMRefiner on each dataset.
An ablation study illustrating the sensitivity of FMRefiner to $\sigma_f$
is reported in Table~\ref{tab:sigma_ablation_fm}.

\paragraph{Limitations.}
\begin{itemize}
    \item Training--inference mismatch: the synthetic $x_1$ distribution used during
    training differs from the upstream generator outputs at inference.
    \item The noise magnitude $\sigma_f$ must be carefully tuned, as mismatches
    propagate directly to the refined samples. In our experiments, we use a
    near-optimal $\sigma_f$ selected from preliminary ablations to ensure stable
    and high-quality refinement.

    \item The refiner is trained along a continuous interpolation path $x_t$,
    making single-step refinement generally infeasible without multi-step integration.
\end{itemize}

\paragraph{Discussion.}
The results highlight the practical challenges of data-space refinement.
Noise-injection methods degrade performance when the injected noise does not align
with the true generative error, even at small noise scales.
FMRefiner further suffers from training--inference mismatch, as the synthetic
perturbations of real data do not accurately reflect the structure of generator errors.
In contrast, our DFR applies mild data augmentation only during training and preserves
the original inference trajectory, leading to consistently improved performance.

\begin{table*}[t]
\centering
\caption{
Ablation study on the perturbation magnitude $\sigma_f$ for FMRefiner.
The intermediate noise scale is fixed to $\sigma_z = 0.1$ for all settings.
Results are reported on CIFAR-10, MNIST, and FFHQ 256$\times$256.
FID $\downarrow$ / sFID $\downarrow$ / IS $\uparrow$ are reported.
}
\label{tab:sigma_ablation_fm}
\small
\begin{tabular}{lccccccccc}
\toprule
$\sigma_f$
& \multicolumn{3}{c}{CIFAR-10}
& \multicolumn{3}{c}{MNIST}
& \multicolumn{3}{c}{FFHQ 256$\times$256} \\
\cmidrule(lr){2-4} \cmidrule(lr){5-7} \cmidrule(lr){8-10}
& FID $\downarrow$ & sFID $\downarrow$ & IS $\uparrow$
& FID $\downarrow$ & sFID $\downarrow$ & IS $\uparrow$
& FID $\downarrow$ & sFID $\downarrow$ & IS $\uparrow$ \\
\midrule
Base
& 4.84 & 0.0034 & 9.04
& 3.95 & 0.0036 & 2.077
&9.47 & 0.0049 & 4.44 \\
0.01
& 4.65 & 0.0033 & 9.08
& 3.21 & 0.0027 & \cellcolor{gray!15}\textbf{2.083}
& 9.45 & 0.0049 & \cellcolor{gray!15}\textbf{4.48} \\
0.05
& \cellcolor{gray!15}\textbf{4.43} & \cellcolor{gray!15}\textbf{0.0026} & \cellcolor{gray!15}\textbf{9.29}
& \cellcolor{gray!15}\textbf{2.87} & \cellcolor{gray!15}\textbf{0.0023} & 2.075
& \cellcolor{gray!15}\textbf{9.27} & 0.0047 & \cellcolor{gray!15}\textbf{4.48} \\
0.10
& 4.50 & 0.0028 & 9.13
& 3.04 & \cellcolor{gray!15}\textbf{0.0023} & 2.062
& 9.35 & \cellcolor{gray!15}\textbf{0.0046} &4.44 \\
0.20
& 4.61 & 0.0031 & 9.24
& 5.25 & 0.0051 & 2.066
& 11.76 & 0.0072 & 4.45 \\
\bottomrule
\end{tabular}
\end{table*}

\section{Discussion on the Choice of Mixing Coefficient $\alpha$}
\label{app:alpha}

In latent-space refinement, the mixing coefficient $\alpha$ controls the strength of stochastic
perturbation applied to the latent correction process, thereby governing the trade-off between
exploration and fidelity.
Importantly, $\alpha$ is not used as a fixed scalar.
Instead, for each sample we draw a random mixing factor
\begin{equation}
    a \sim \mathcal{U}(0, \alpha),
\end{equation}
and apply it as
\begin{equation}
    a = \alpha \cdot \mathrm{rand}(\cdot),
\end{equation}
where $\mathrm{rand}(\cdot)$ denotes element-wise sampling from a uniform distribution on $[0,1]$.
Thus, $\alpha$ defines the \emph{maximum} perturbation strength, while the actual mixing varies
across samples.

Larger values of $\alpha$ increase the diversity of latent perturbations and encourage broader
exploration of the latent space, which can help correct systematic bias.
However, excessive perturbation may disrupt semantic consistency and degrade sample fidelity.
Conversely, smaller $\alpha$ restricts exploration and may limit the effectiveness of refinement.

\paragraph{Ablation Results.}
Table~\ref{tab:alpha_ablation} reports an ablation study over different choices of $\alpha$ on
CIFAR-10 and MNIST.
We observe that moderate values of $\alpha$ consistently yield the best performance across datasets.
In particular, $\alpha = 0.2$ achieves the strongest overall improvement on CIFAR-10,
while $\alpha = 0.1$ performs best on MNIST, leading to the lowest FID and sFID.
Both overly small and overly large values of $\alpha$ result in degraded performance,
highlighting the importance of balanced stochastic mixing.

These results indicate that latent-space refinement benefits from controlled randomness rather than
deterministic or overly aggressive perturbations.
Based on this study, we fix $\alpha = 0.2$ for CIFAR-10 and $\alpha = 0.1$ for MNIST
in all other experiments.

\begin{table*}[t]
\centering
\caption{
Ablation study on the mixing coefficient $\alpha$ for latent-space refinement.
Results are reported on CIFAR-10, MNIST, and FFHQ 256$\times$256.
FID $\downarrow$ / sFID $\downarrow$ / IS $\uparrow$ are reported.
}
\label{tab:alpha_ablation}
\small
\begin{tabular}{lccccccccc}
\toprule
 $\alpha$& \multicolumn{3}{c}{CIFAR-10} & \multicolumn{3}{c}{MNIST} & \multicolumn{3}{c}{FFHQ 256$\times$256} \\
 \cmidrule(lr){2-4} \cmidrule(lr){5-7} \cmidrule(lr){8-10}
& FID $\downarrow$ & sFID $\downarrow$ & IS $\uparrow$
& FID $\downarrow$ & sFID $\downarrow$ & IS $\uparrow$
& FID $\downarrow$ & sFID $\downarrow$ & IS $\uparrow$\\
\midrule
Base  &4.84 & 0.0034 & 9.04 & 3.95 & 0.0036 & 2.077& 9.47 & 0.0049 & \cellcolor{gray!15}\textbf{4.44} \\
0.0   & 4.16 & 0.0022 & 9.32 & 2.02 & 0.0012 & 2.109& 8.85& \cellcolor{gray!15}\textbf{0.0040}& 4.29 \\
0.1   & 4.24 & 0.0022 & 9.29 & \cellcolor{gray!15}\textbf{1.46} & \cellcolor{gray!15}\textbf{0.0008} & 2.085& 8.78& \cellcolor{gray!15}\textbf{0.0040}& 4.36 \\
0.2   & \cellcolor{gray!15}\textbf{3.10} & \cellcolor{gray!15}\textbf{0.0014} & \cellcolor{gray!15}\textbf{9.41} & 2.33 & 0.0016 & 2.115&\cellcolor{gray!15}\textbf{8.75} & 0.0041 & 4.41 \\
0.3   & 4.20 & 0.0022 & 9.36 & 2.05 & 0.0011 & \cellcolor{gray!15}\textbf{2.120}& 9.05& 0.0043& 4.37 \\
0.5   & 4.20 & 0.0022 & 9.27 & 2.65 & 0.0017 & 2.104& 9.07& 0.0043& 4.36 \\
\bottomrule
\end{tabular}
\end{table*}

\section{Additional Results: Molecular Generation}
\label{app:molecular}

We further evaluate the proposed Bi-stage Flow Refinement (BFR) framework on molecular conformer generation tasks, following standard benchmarks used in prior work.
Experiments are conducted on the small peptide ALA2 and the larger protein Chignolin Mutant.
Base generative models are diffusion-based models (DDPM or FM), combined with existing data-space refiners and our proposed BFR refiners.
\begin{figure}[t]
    \centering
    \begin{subfigure}{0.48\linewidth}
        \centering
        \includegraphics[width=\linewidth]{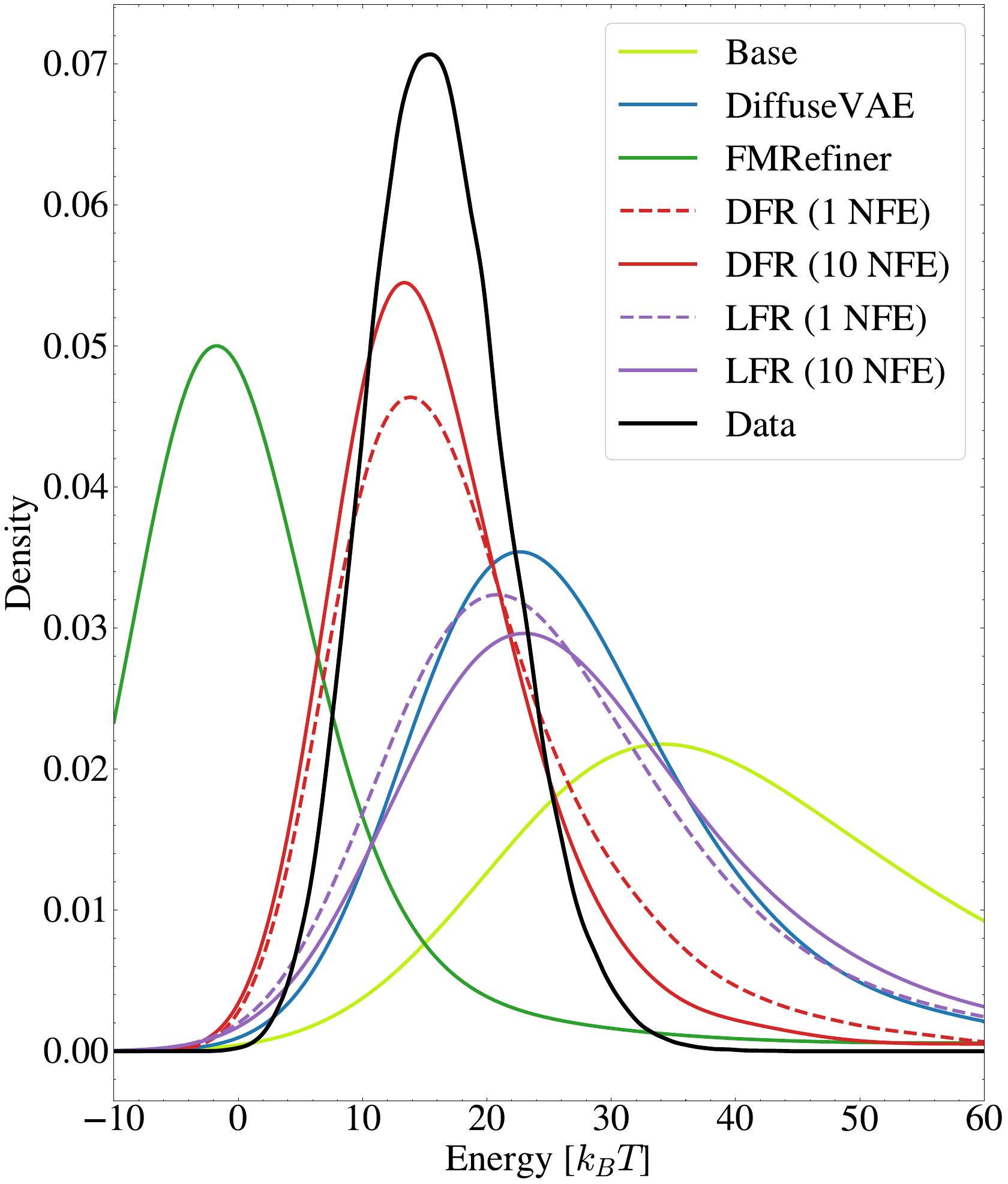}
        \caption{ALA2}
    \end{subfigure}
    \begin{subfigure}{0.48\linewidth}
        \centering
        \includegraphics[width=\linewidth]{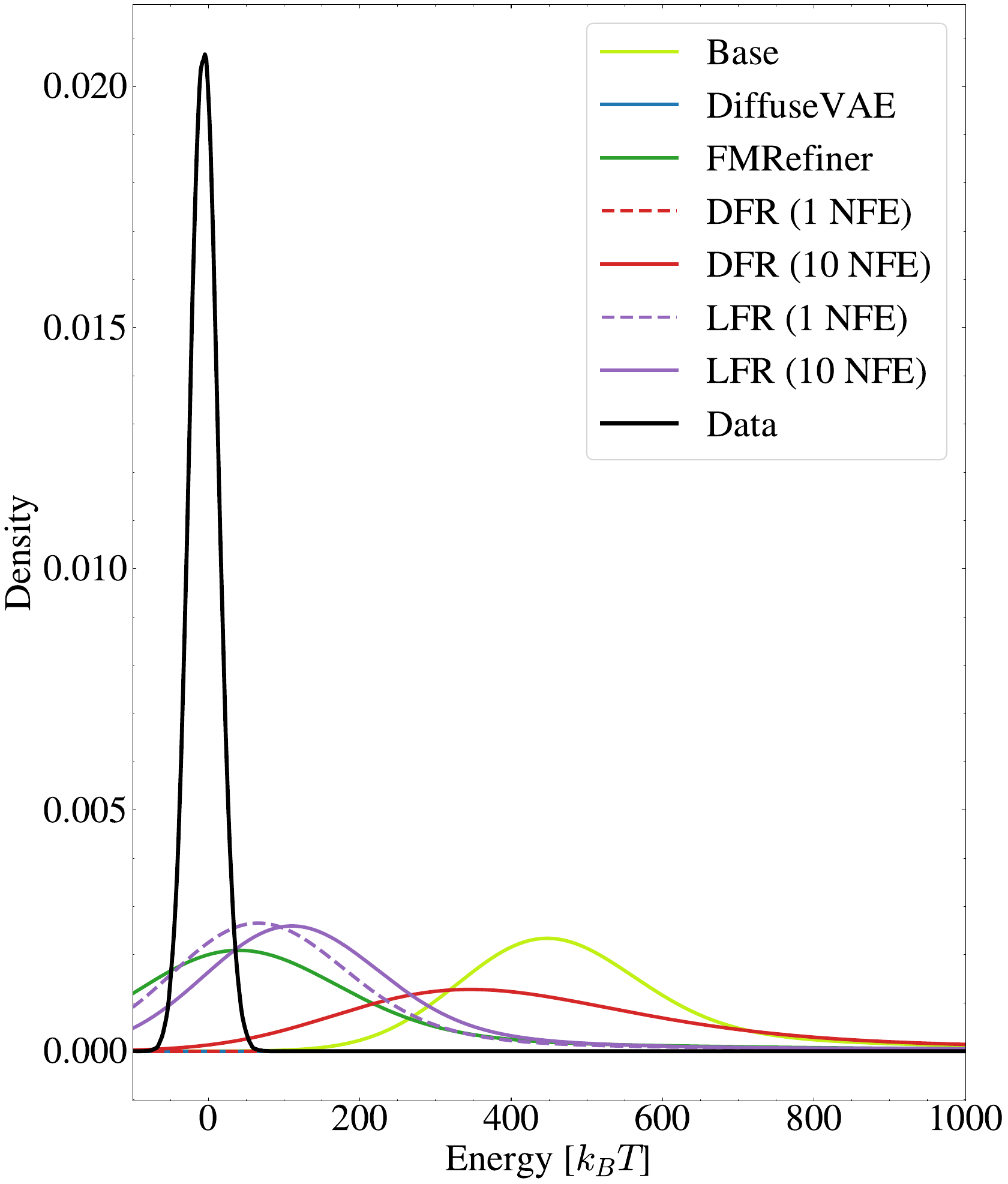}
        \caption{Chignolin Mutant}
    \end{subfigure}
    \caption{Energy distribution comparison.
    Energy distributions of samples generated by the base model, existing data-space refiners,
    and BFR variants, shown alongside the reference data distribution.}
    \label{fig:energy_distribution_comparison}
\end{figure}

\paragraph{Evaluation metrics.}
We report Coverage (COV) and Average Minimum RMSD (AMR) in both Recall and Precision variants.
COV measures the diversity of generated conformations relative to reference ensembles, while AMR quantifies geometric accuracy by computing the minimum RMSD between generated and reference conformers.
All metrics are evaluated using the same number of generated samples (10K) for all methods to ensure a fair comparison.
For existing refinement baselines, DiffuseVAE~\cite{pandey2022diffusevae} and FMRefiner~\cite{xu2025flow} are both evaluated with 10 NFEs.
In addition, we assess energy-based quality by measuring the relative difference (in percentage) between the average energy of generated samples and that of the reference data distribution.

Let $\mathcal{X}_{\mathrm{ref}} = \{x_i\}_{i=1}^{N}$ denote the reference conformations and 
$\mathcal{X}_{\mathrm{gen}} = \{y_j\}_{j=1}^{M}$ the generated samples.
Let $\mathrm{RMSD}(x,y)$ denote the root-mean-square deviation after optimal alignment, and $\tau$ be a predefined RMSD threshold.

\textbf{Coverage (Recall)} is defined as
\begin{equation}
\mathrm{COV}_{\mathrm{R}} =
\frac{1}{N}
\sum_{i=1}^{N}
\mathbf{1}
\left(
\min_{j} \mathrm{RMSD}(x_i, y_j) < \tau
\right),
\end{equation}
measuring the fraction of reference conformations that are covered by the generated set.

\textbf{Coverage (Precision)} is defined as
\begin{equation}
\mathrm{COV}_{\mathrm{P}} =
\frac{1}{M}
\sum_{j=1}^{M}
\mathbf{1}
\left(
\min_{i} \mathrm{RMSD}(y_j, x_i) < \tau
\right),
\end{equation}
measuring the fraction of generated samples that are close to the reference ensemble.

\textbf{Average Minimum RMSD (Recall)} is defined as
\begin{equation}
\mathrm{AMR}_{\mathrm{R}} =
\frac{1}{N}
\sum_{i=1}^{N}
\min_{j} \mathrm{RMSD}(x_i, y_j),
\end{equation}
while \textbf{Average Minimum RMSD (Precision)} is
\begin{equation}
\mathrm{AMR}_{\mathrm{P}} =
\frac{1}{M}
\sum_{j=1}^{M}
\min_{i} \mathrm{RMSD}(y_j, x_i).
\end{equation}

Finally, the \textbf{energy error} is defined as the signed relative difference between
the average energies of generated and reference samples,
\begin{equation}
\mathrm{Energy}\;\% =
\frac{
\mathbb{E}_{y \sim \mathcal{X}_{\mathrm{gen}}}[E(y)]
-
\mathbb{E}_{x \sim \mathcal{X}_{\mathrm{ref}}}[E(x)]
}{
\left|
\mathbb{E}_{x \sim \mathcal{X}_{\mathrm{ref}}}[E(x)]
\right|
}
\times 100.
\end{equation}

\begin{table*}[t]
\centering
\caption{Molecular generation results on ALA2 and Chignolin Mutant. 
Metrics: Coverage (COV) and Average Minimum RMSD (AMR) in Recall/Precision, and energy error (\%). 
COV and AMR are computed using relative tolerance $\tau=0.05\,\text{\AA}$ for ALA2 and $\tau=0.35\,\text{\AA}$ for Chignolin Mutant.
Base models: DDPM or FM, combined with existing refiners and BFR.}
\label{tab:mol_results_twocol}
\small
\setlength{\tabcolsep}{3pt}
\begin{tabular}{l@{\hskip 2pt}cccccccccc@{\hskip 2pt}}
\toprule
& \multicolumn{5}{c}{ALA2} & \multicolumn{5}{c}{Chignolin Mutant} \\
\cmidrule(lr){2-6} \cmidrule(lr){7-11}
Model 
& COV$_\text{R}$ $\uparrow$ & COV$_\text{P}$ $\uparrow$ 
& AMR$_\text{R}$ $\downarrow$ & AMR$_\text{P}$ $\downarrow$ 
& Energy $\%$ $\downarrow$
& COV$_\text{R}$ $\uparrow$ & COV$_\text{P}$ $\uparrow$ 
& AMR$_\text{R}$ $\downarrow$ & AMR$_\text{P}$ $\downarrow$ 
& Energy $\%$ $\downarrow$ \\
\midrule
Base & 62.25 & 81.51 & 0.049 & 0.043 & 212.25 & 32.17 & 69.45 & 0.387 & 0.317 & 13545 \\
+ DiffuseVAE~\cite{pandey2022diffusevae}  & 64.88 & 85.54 & 0.048 & 0.042 & 90.13 & 25.22 & 52.09 & 0.397 & 0.343 & 34297\\
+ FMRefiner~\cite{xu2025flow}  & 69.65 & 85.49 & 0.047 & 0.040 & -60.80 & 32.98 & 69.73 & 0.386 & 0.315 & 7756\\
\midrule
+ DFR (1-NFE) & 65.54 & 86.77 & 0.048 & 0.041 & 22.78 & 0.01 & 0.08 & 0.519 & 0.462 & 34297 \\
+ DFR (10-NFE)& 65.75 & 85.82 & 0.048 & 0.041 & 9.68 & 32.92 & 69.53 & 0.386 & 0.315 & 15630 \\
+ LFR (1-NFE) & 65.16 & 86.59 & 0.048 & 0.041 & 95.43 & 33.10 & 70.61 & 0.386 & 0.314 & 6387 \\
+ LFR (10-NFE)& 64.08 & 84.87 & 0.048 & 0.042 & 117.27 & 32.86 & 70.74 & 0.386 & 0.314 & 7277 \\
\bottomrule
\end{tabular}
\end{table*}

\paragraph{Results on ALA2.}
Quantitative results on ALA2 are summarized in Table~\ref{tab:mol_results_twocol}.
Compared to the base model, both existing refiners and BFR variants improve conformational
coverage and geometric accuracy.
Beyond structural metrics, Figure~\ref{fig:energy_distribution_comparison} compares the energy
distributions of generated samples against the reference ensemble.
The base model exhibits a pronounced energy mismatch, while refinement consistently shifts
the distribution toward the target.
In particular, BFR substantially reduces the energy discrepancy, indicating improved
consistency with the underlying physical energy landscape.

\paragraph{Results on Chignolin Mutant.}  
Due to the presence of extreme outliers in the molecular energies of the Chignolin Mutant dataset
(with values exceeding $10^{10}$), we apply energy truncation when computing this metric.
Specifically, only samples with energies $E < 2000$ are included when estimating the expectation,
and the energy error is computed on this truncated set. The number of excluded samples per method is:
Base=1742, 
DiffuseVAE=10000, 
FMRefiner=1660, 
DFR (1-NFE)=10000, 
DFR (10-NFE)=26692, 
LFR (1-NFE)=1236, 
LFR (10-NFE)=1352.
These counts are also reported in the accompanying repository for full transparency.

Table~\ref{tab:mol_results_twocol} reports results on the challenging Chignolin Mutant system, where all methods show lower conformational coverage and higher RMSD than ALA2. Notably, DFR (1-NFE) exhibits a critical failure mode, with COV$_\text{R}$ dropping from 32.17 to 0.01; this arises because the base model already performs poorly, and DFR relies on relatively large-strength data augmentation, so very few function evaluations cannot adequately fit the target distribution. DiffuseVAE also shows degradation in coverage and geometric accuracy. From an energy perspective, the base model exhibits a severe mismatch with the target distribution. FMRefiner substantially reduces the energy error (from 13545\% to 7756\%), while LFR further achieves a markedly larger and more stable reduction (e.g., LFR 1-NFE decreases the error to 6387\%, roughly halving it relative to the base model). Although DFR (10-NFE) increases the average energy error overall, Figure~\ref{fig:energy_distribution_comparison} shows that it successfully corrects a subset of samples toward lower-energy regions, indicating partial recovery despite overall instability. 

\paragraph{Discussion.}
While different refinement strategies exhibit varying trade-offs between coverage, accuracy, and energy alignment, the results demonstrate that post-hoc refinement can effectively improve molecular generation without retraining the base model.
These findings complement the main image-generation experiments and further support the generality of BFR across domains with structured and physically grounded target distributions.

\section{Conditional vs. Unconditional CIFAR-10 Results}
\label{app:cifar_cond_uncond}
\begin{table*}[t]
\centering
\caption{
Comparison of conditional and unconditional image generation results on CIFAR-10.
FID $\downarrow$ / sFID $\downarrow$ / IS $\uparrow$ are reported.
}
\label{tab:cifar_cond_uncond}
\small
\begin{tabular}{lcccccc}
\toprule
 & \multicolumn{3}{c}{Unconditional} & \multicolumn{3}{c}{Conditional} \\
\cmidrule(lr){2-4} \cmidrule(lr){5-7}
Model & FID $\downarrow$ & sFID $\downarrow$ & IS $\uparrow$
      & FID $\downarrow$ & sFID $\downarrow$ & IS $\uparrow$ \\
\midrule
Base & 3.77 & 0.0018 & 9.04 & 4.84 & 0.0034 & 9.04 \\
+ DFR  & 3.54 & 0.0013 & 9.39 & 3.96 & 0.0020 & 9.40 \\
+ LFR  & 3.36 & 0.0010 & 9.40 & 3.10 & 0.0014 & 9.41 \\
\bottomrule
\end{tabular}
\end{table*}
\paragraph{Discussion.}
For completeness, we report both unconditional and conditional CIFAR-10 results in Table~\ref{tab:cifar_cond_uncond}.
Both settings exhibit consistent qualitative trends: data-space and latent-space refinement systematically improve upon the base generator.
Quantitatively, conditional models achieve stronger performance across all metrics, and therefore serve as the default configuration throughout the main paper.
\section{Robustness of BFR Across Base Models}
\label{Robustness}

In this appendix, we provide a more detailed analysis of the base-model transferability of BFR refiners on the MNIST dataset.

\subsection{FID Performance}
Table~\ref{tab:robustness_fid} summarizes the FID scores for both refinement strategies under the transfer setting.

\begin{table}[h!]
\centering
\caption{FID of BFR refiners under base-model transfer on MNIST.}
\label{tab:robustness_fid}
\begin{tabular}{lccc}
\toprule
Model & FID $\downarrow$ & sFID $\downarrow$ & IS $\uparrow$ \\
\midrule
Base & 7.61 & 0.0053 & 2.084 \\
+ DFR (1-NFE) & 5.04 & 0.0037 & \textbf{2.141} \\
+ DFR (10-NFE) & \textbf{4.62} & \textbf{0.0032} & 2.121 \\
+ LFR (1-NFE) & 6.86 & 0.0058 & 2.101 \\
+ LFR (10-NFE) & 8.64 & 0.0076 & 2.146 \\
\bottomrule
\end{tabular}
\end{table}
As shown, DFR achieves a larger reduction in FID compared to LFR, suggesting that direct correction in the data space is more effective for mitigating systematic generative bias. Nevertheless, both refiners significantly improve sample quality relative to the degraded base model, confirming the robustness of BFR to variations in generator quality.
\begin{figure}[t]
    \centering
    \includegraphics[width=\linewidth]{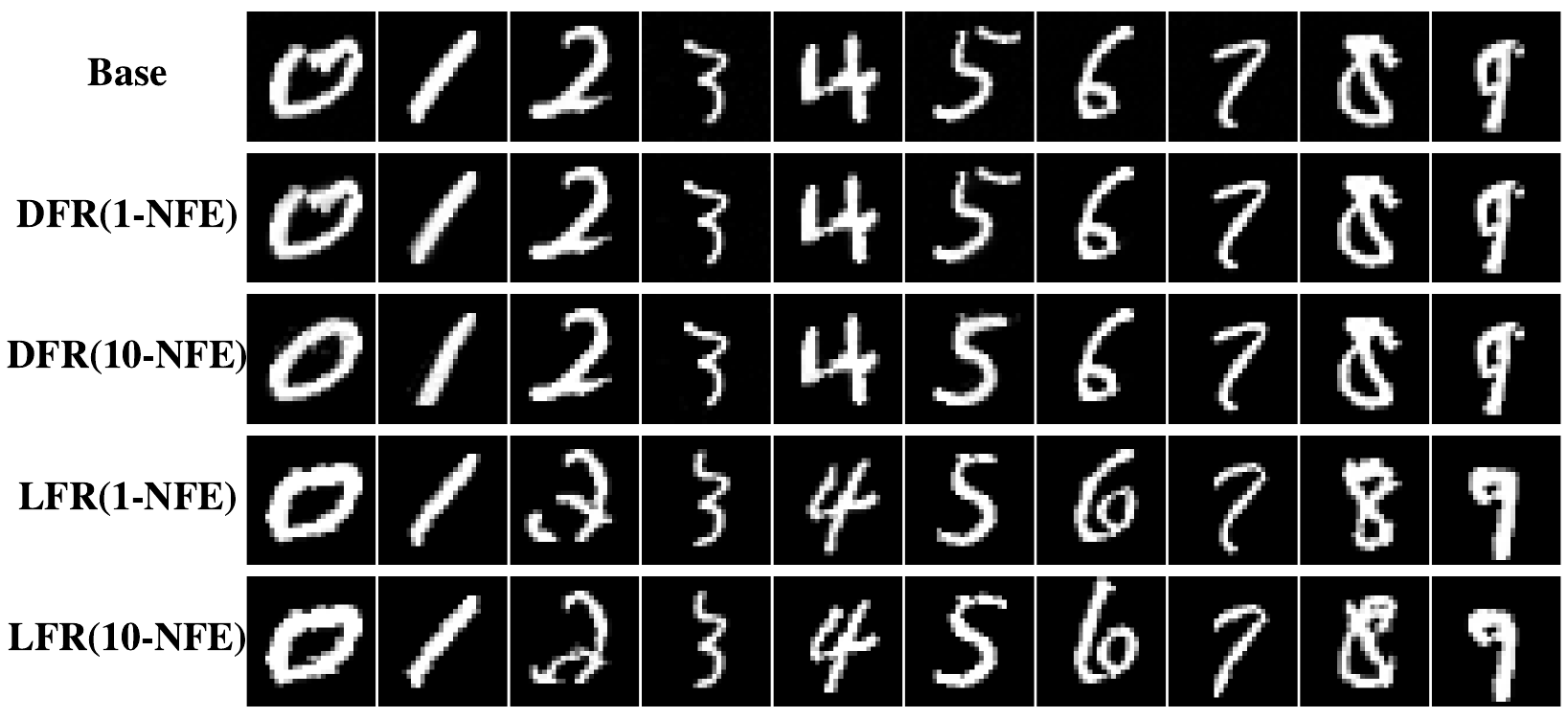}
    \caption{Qualitative comparison of MNIST samples before and after refinement.
Each row corresponds to a refinement method, and samples are shown without random selection.
The base generator produces samples with an FID of 7.61.
While most BFR variants lead to noticeable improvements in visual quality over the base model,
this improvement is not universal: \textbf{LFR (10-NFE)} exhibits a degradation in perceptual quality.
In contrast, \textbf{DFR (10-NFE)} achieves the highest perceptual quality among all methods,
which is consistent with its superior FID performance.
}
    \label{fig:mnist_refinement_qualitative2}
\end{figure}
\subsection{Experimental Setup}
We consider two BFR variants: data-space refinement (DFR) and latent-space flow refinement (LFR). Both models are trained on a base generator achieving an FID of 3.9. We then evaluate their performance when applied to a different base generator whose FID degrades to 7.6, without any retraining.
\subsection{Discussion}
These results demonstrate that BFR can capture transferable patterns of
generative bias and be reused across multiple base generators, reducing
training cost and simplifying deployment. While most refiners improve sample
quality, the effect can depend on the number of function evaluations (NFE) and
the specific refiner: for instance, LFR improves FID with 1-NFE but
degrades FID with 10-NFE on MNIST. This behavior is consistent with other
experiments where LFR with fewer NFEs often performs better than with more
NFEs. The likely reason is over-refinement: excessive iterations may push the
samples too far, reducing fidelity to the base distribution, thereby worsening
FID.

\section{Generator and Inverse Implementation with Sampling Cost}

LFR and DFR are trained using the same flow-matching procedure as their
base generators (DDPM and FM). During sampling, generator inversion (mapping
data to latent space) is realized via backward integration along the
probability flow ODE. The training cost of LFR and DFR is comparable to the
base models.  
Following DiffuseVAE~\cite{pandey2022diffusevae} and FMRefiner~\cite{xu2025flow}, 
we use 10 function evaluations (NFE) for ODE sampling.  

In addition, DFR training employs lightweight data augmentation to improve
stability, including Gaussian noise and slight blurring, with specific types
and hyperparameters (e.g., noise magnitude and blur kernel size) provided in
the accompanying repository. LFR training details, including any data
preprocessing or augmentation strategies, as well as all additional
implementation details, solver settings, and reproducibility instructions, are
also fully documented at
\url{https://github.com/XinPeng76/Rethinking-Refinement}.

\section{Statement on the Use of Large Language Models}
\label{app:llm_statement}

During the preparation of this manuscript, large language models (LLMs) were used in a limited manner
solely for language editing purposes, such as improving clarity, grammar, and academic style.
All aspects of the research conception, methodological development, experimental design,
analysis of results, and the scientific conclusions presented in this paper
were carried out independently by the authors.


\end{document}